\documentclass[letterpaper, 10 pt, journal, twoside]{IEEEtran}

\usepackage{amsmath}
\usepackage{amssymb}
\usepackage{bbm}
\usepackage{booktabs}
\usepackage{comment}
\usepackage{dblfloatfix}
\usepackage{graphicx} 
\usepackage{multirow}
\usepackage{microtype}
\usepackage{siunitx}

\usepackage[nobreak]{cite}

\usepackage[inline]{enumitem}
\newcommand{\enumlabel}{\textbf{(\roman*)}}

\usepackage{array}
\newcolumntype{M}[1]{>{\centering\arraybackslash}m{#1}}

\usepackage{subcaption}
\usepackage[breaklinks, hidelinks]{hyperref}  

\usepackage[capitalize]{cleveref}
\crefname{section}{Sec.}{Secs.}
\Crefname{section}{Section}{Sections}
\Crefname{table}{Table}{Tables}
\crefname{table}{Tab.}{Tabs.}

\newcommand{\bb}{\boldsymbol}  
\newcommand{\tb}{\textbf}  

\newcommand{\norm}[1]{\left\lVert#1\right\rVert}

\title{
SyMFM6D: Symmetry-aware Multi-directional Fusion for Multi-View 6D Object Pose Estimation
}

\author{
Fabian Duffhauss$^{1, 2}$, 
Sebastian Koch$^{1, 3}$, 
Hanna Ziesche$^{1}$, 
Ngo Anh Vien$^{1}$, and
Gerhard Neumann$^{4}$% 
\thanks{$^{1}$Fabian Duffhauss, Sebastian Koch, Hanna Ziesche, and Ngo Anh Vien are with the Bosch Center for Artificial Intelligence, Renningen, Germany
        {\tt\footnotesize Fabian.Duffhauss@bosch.com}}%
\thanks{$^{2}$Fabian Duffhauss is with the University of Tübingen, Tübingen, Germany}%
\thanks{$^{3}$Sebastian Koch is with the Ulm University, Ulm, Germany}%
\thanks{$^{4}$Gerhard Neumann is with the Institute for Anthropomatics and Robotics, Karlsruhe Institute of Technology, Karlsruhe, Germany
        {\tt\footnotesize Gerhard.Neumann@kit.edu}}%
\thanks{Source code, datasets, and implementation details are available at \url{https://github.com/boschresearch/SyMFM6D}.}%
}

\begin{document}
\bstctlcite{IEEEexample:BSTcontrol}

\maketitle

\begin{abstract}

Detecting objects and estimating their 6D poses is essential for automated systems to interact safely with the environment. Most 6D pose estimators, however, rely on a single camera frame and suffer from occlusions and ambiguities due to object symmetries. 
We overcome this issue by presenting a novel symmetry-aware multi-view 6D pose estimator called SyMFM6D. Our approach efficiently fuses the RGB-D frames from multiple perspectives in a deep multi-directional fusion network and predicts predefined keypoints for all objects in the scene simultaneously. Based on the keypoints and an instance semantic segmentation, we efficiently compute the 6D poses by least-squares fitting. To address the ambiguity issues for symmetric objects, we propose a novel training procedure for symmetry-aware keypoint detection including a new objective function. 
Our SyMFM6D network significantly outperforms the state-of-the-art in both single-view and multi-view 6D pose estimation. We furthermore show the effectiveness of our symmetry-aware training procedure and demonstrate that our approach is robust towards inaccurate camera calibration and dynamic camera setups.

\end{abstract}
\section{Introduction}

Estimating the 6D poses of objects is an essential computer vision task which is widely used in robotics \cite{posecnn, pvn3d, ffb6d}, automated driving \cite{avod, gu2021ecpc_icp}, and augmented reality \cite{arSurvey, su2019ar}.
In recent years, 6D pose estimators have made significant progress based on deep neural network architectures which rely on a single RGB image \cite{posecnn, so_pose, zebrapose}, on a single point cloud \cite{pointvotenet2020, votingAttentionPoseEst22},
or fuse both \cite{densefusion, pvn3d, ffb6d}. 
Single-view methods, however, have problems detecting objects which are occluded by other objects.
These problems can be overcome by considering data from multiple perspectives.
Fusing multi-view data can significantly improve the accuracy and robustness of environmental understanding in complex scenarios, which can enable more flexible production and assembly processes, among other applications.
There are already a few methods that consider multi-view data \cite{zeng2017multi, li2018unified, cosypose} which are, however, computationally expensive and not designed for scenes with strong occlusions. Moreover, most methods suffer from symmetric objects as they have multiple 6D poses with same visual and geometric appearance, causing most learning-based estimators to average over these multiple solutions.

\begin{figure}[t]
  \centering 
  \includegraphics[page=1, trim = 10mm 35mm 7mm 36mm, clip, width=1.0\columnwidth]{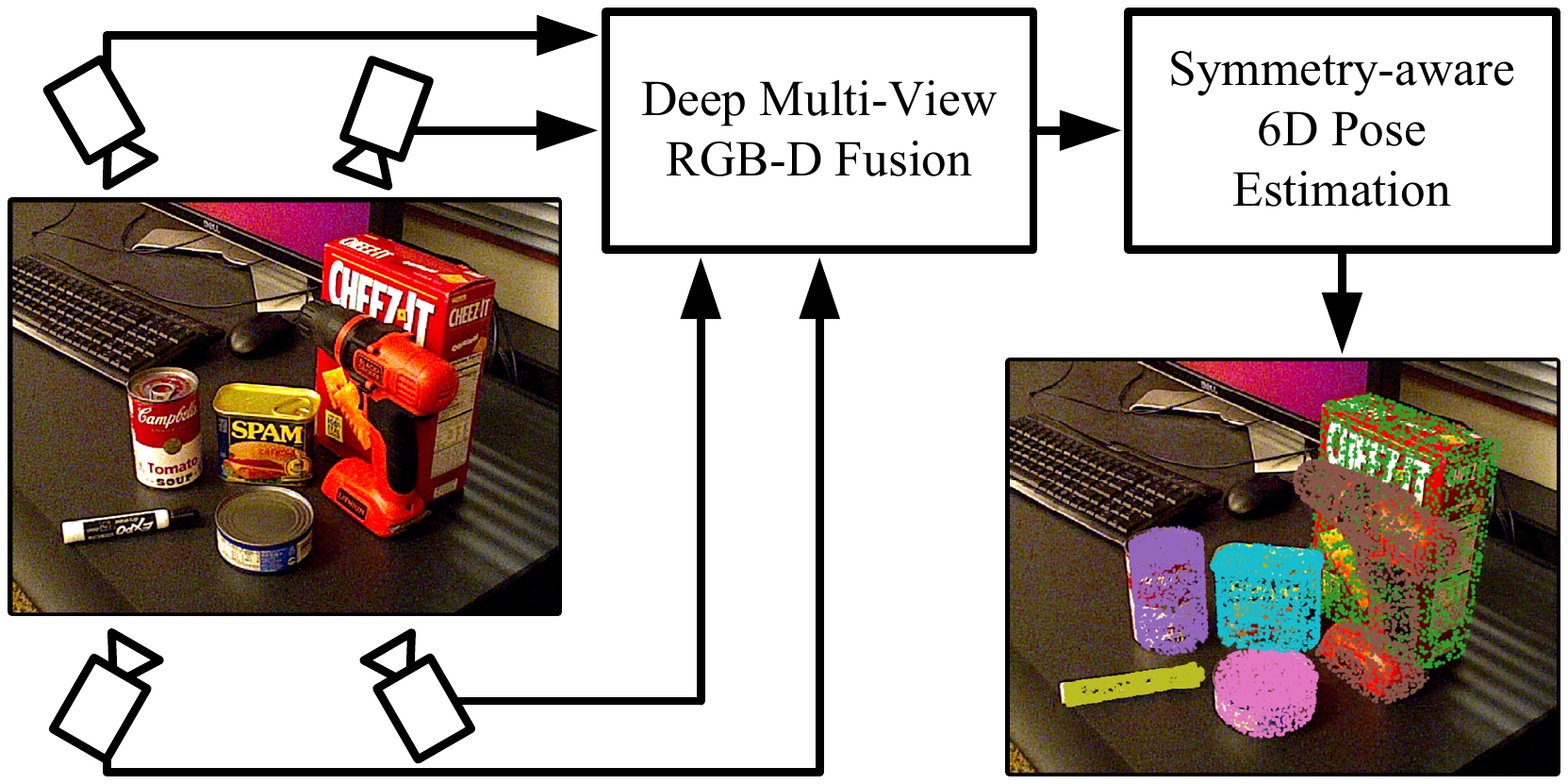}
   \caption{Overview of our proposed SyMFM6D approach. We present a novel deep multi-directional fusion network which merges RGB-D data from multiple cameras. SyMFM6D predicts the 6D poses of all objects in the scene while considering object symmetries. It copes with very cluttered scenes and outperforms the state-of-the-art in single-view and multi-view 6D pose estimation.}
   \label{fig_eye_catcher}
\end{figure}

We present a novel \textbf{Sy}mmetry-aware \textbf{M}ulti-directional \textbf{F}usion approach for \textbf{M}ulti-view \textbf{6D} pose estimation called SyMFM6D which overcomes the previously mentioned issues. \cref{fig_eye_catcher} shows an overview of our system. 
SyMFM6D exploits the visual and geometric information from an arbitrary number of RGB-D images depicting a scene from multiple perspectives. We propose a deep multi-directional fusion network which fuses the multi-view RGB-D data efficiently and learns a compact representation of the entire scene. Our approach predicts the 6D poses of all objects in the scene simultaneously based on keypoint detection, instance semantic segmentation, and least-squares fitting. Furthermore, we present a novel symmetry-aware training procedure including a novel objective function which significantly improves the keypoint detection. 

Our experiments demonstrate a large benefit of the proposed symmetry-aware training procedure, improving the accuracy of both symmetric and non-symmetric objects due to synergy effects. Thus, our approach outperforms the state-of-the-art in single-view 6D pose estimation. SyMFM6D also outperforms the state-of-the-art multi-view approach while being computationally more efficient. We furthermore show that our approach works accurately in both fixed and dynamic camera settings. Moreover, our method is robust towards inaccurate camera calibration by compensating imprecise camera pose information when using multiple views.

Our main contributions are:

\begin{enumerate*}[label=\enumlabel]
    \item We propose a novel multi-directional multi-view fusion network for efficient representation learning of multiple RGB-D frames and present a novel multi-view 6D pose estimation method based on it.
    \item We present a novel symmetry-aware training procedure for 3D keypoint detection based on a symmetry-aware objective function.
    \item We present a novel synthetic dataset with photorealistic multi-view \mbox{RGB-D} data and labels for 6D pose estimation as well as instance semantic segmentation.
	\item We demonstrate significant improvements and synergy effects due to our symmetry-aware training procedure on challenging datasets including symmetric and non-symmetric objects.
	\item Our method outperforms the state-of-the-art in single-view and multi-view 6D object pose estimation. We further demonstrate the robustness of our approach towards inaccurate camera calibration and dynamic camera setups. 
\end{enumerate*}

\section{Related Work}

Over the last few years, there has been significant progress in the area of 6D pose estimation. We now discuss the most important milestones subdivided into single-view methods, multi-view methods, and symmetry-aware methods.

\subsection{Single-View 6D Pose Estimation}

The methods in this family require only a single input modality, which can be RGB, point cloud, or RGB-D. Traditional pose estimators using a \emph{single RGB image} are mostly feature-based \cite{lowe1999sift, lowe2004distinctive, rosten2006machine, moped, collet2009object, pvnet} or based on template matching \cite{ComparingImgsTPAMI93, gu2010discriminative, 2011gradientResponseMaps, cao2016realTime6d}. Especially the former group of methods are often multi-staged and first extract local features from the given RGB image before matching the 2D-3D-correspondences to estimate the object's pose using a Perspective-n-Point (PnP) algorithm \cite{ransac}. End-to-end trainable neural networks directly predict object poses without requiring multiple stages \cite{2015viewpointsKeypoints, deepim, posecnn, gupta2019cullnet, ssd6d, tekin2018real, gdrnet, so_pose, zebrapose}. 
These methods share similar ideas to exploit differentiable PnP or differentiable rendering techniques.

The recent advance of LiDAR and depth sensors promoted the proposal of methods based on a \emph{single point cloud} \cite{chen2020survey6d, fernandes2021pointCloudSurvey}. These methods apply either 3D convolutions \cite{song2014sliding, li2017_3dFCN}, or variations of PointNet \cite{pointNet} as backbone \cite{voxelnet, second, pointpillars}. The authors of \cite{qi2019voteNet} and \cite{xie2021VENet} introduce and further improve voting techniques for 3D object detection.
However, since point cloud based methods cannot extract texture information, their application range is limited.

In contrast, \emph{RGB-D based approaches} can combine the advantages of both modalities. For instance, \cite{avod} and \cite{mv3d} fuse an RGB image with a LiDAR point cloud by applying networks for convolutional feature extraction and for generating 3D object proposals.
The approaches proposed in \cite{pointfusion}, \cite{densefusion}, and \cite{pvn3d} separately process the RGB image by a CNN and the point cloud by a PointNet-based network before fusing the appearance features and the geometric features with a dense fusion network.
In \cite{pvn3d} and \cite{ffb6d} the authors employ a deep Hough voting network for 3D keypoint detection before estimating 6D poses by least-squares fitting \cite{leastSquares}. 
However, most previous methods do not consider object symmetries and suffer from strong occlusions.

\subsection{Multi-View 6D Pose Estimation}

Multi-view pose estimators consider multiple RGB(-D) frames showing the same scene from different perspectives in order to reduce the effect of occlusions and to improve the 6D pose estimation accuracy.
The approach proposed in \cite{zeng2017multi} first segments all frames with a CNN before aligning the known 3D models with the segmented object point cloud to estimate their poses.
The authors of \cite{li2018unified} present an end-to-end trainable CNN-based architecture based on a single RGB or RGB-D image. They perform the single-view pose estimation multiple times with images from different viewpoints before selecting the best hypothesis using a voting score that suppresses outliers. 
In \cite{cosypose} the authors propose a three-stage approach which first employs a CNN for generating object candidate proposals for each view independently. Secondly, they conduct a candidate matching considering the predictions of all views before finally performing a refinement procedure based on object-level bundle adjustment \cite{triggs2000bundle}.
The approach of \cite{mv6d} directly fuses the features from multiple RGB-D views before predicting the poses based on keypoints and least-squares fitting \cite{leastSquares}. However, the method uses a computationally expensive feature extraction and fusion network which does not consider object symmetries and it is evaluated on only synthetic datasets.

\subsection{Symmetry-aware 6D Pose Estimation}

Symmetric objects are known to be a challenge for 6D pose estimation approaches due to ambiguities \cite{objectSym6D_3DV19}. Different techniques have been proposed to address this issue. The authors of \cite{objectSym6D_3DV19} and \cite{rad2017bb8} propose to utilize an additional output channel to classify the type of symmetry and its domain range. 
In \cite{pix2poseICCV19}, a loss is introduced that is the smallest error among symmetric pose proposals in a finite pool of symmetric poses. 
In \cite{eposCVPR20} the authors propose to use compact surface fragments as a compositional way to represent objects. As a result, this representation can easily allow handling of symmetries. 
The authors of \cite{zhang2020symmetry6d} employ an additional symmetry prediction as output, and an extra refining step of predicted symmetry via an optimization function. 
A novel output space representation for CNNs is presented in  \cite{symCnnICRA21} where symmetrical equivalent poses are mapped to the same values. In \cite{es6d} the authors introduce a compact shape representation based on grouped primitives to handle symmetries.
However, non of these methods outperforms the keypoint-based methods 
\cite{pvn3d} and \cite{ffb6d}, even though they do not consider object symmetries. In contrast, our method extends current keypoint based methods to consider object symmetries, and consequently outperforms all previous methods on single and multi-view scenes.

\section{Proposed Method: SyMFM6D}

We propose a deep multi-directional fusion approach called SyMFM6D that estimates the 6D object poses of all objects in a cluttered scene based on multiple RGB-D images while considering object symmetries. 
In this section, we define the task of multi-view 6D object pose estimation and present our multi-view deep fusion architecture.

\begin{figure*}[tbh]
  \vspace{2mm}
  \centering
  \includegraphics[page=1, trim = 5mm 40mm 5mm 42mm, clip,  width=1.0\linewidth]{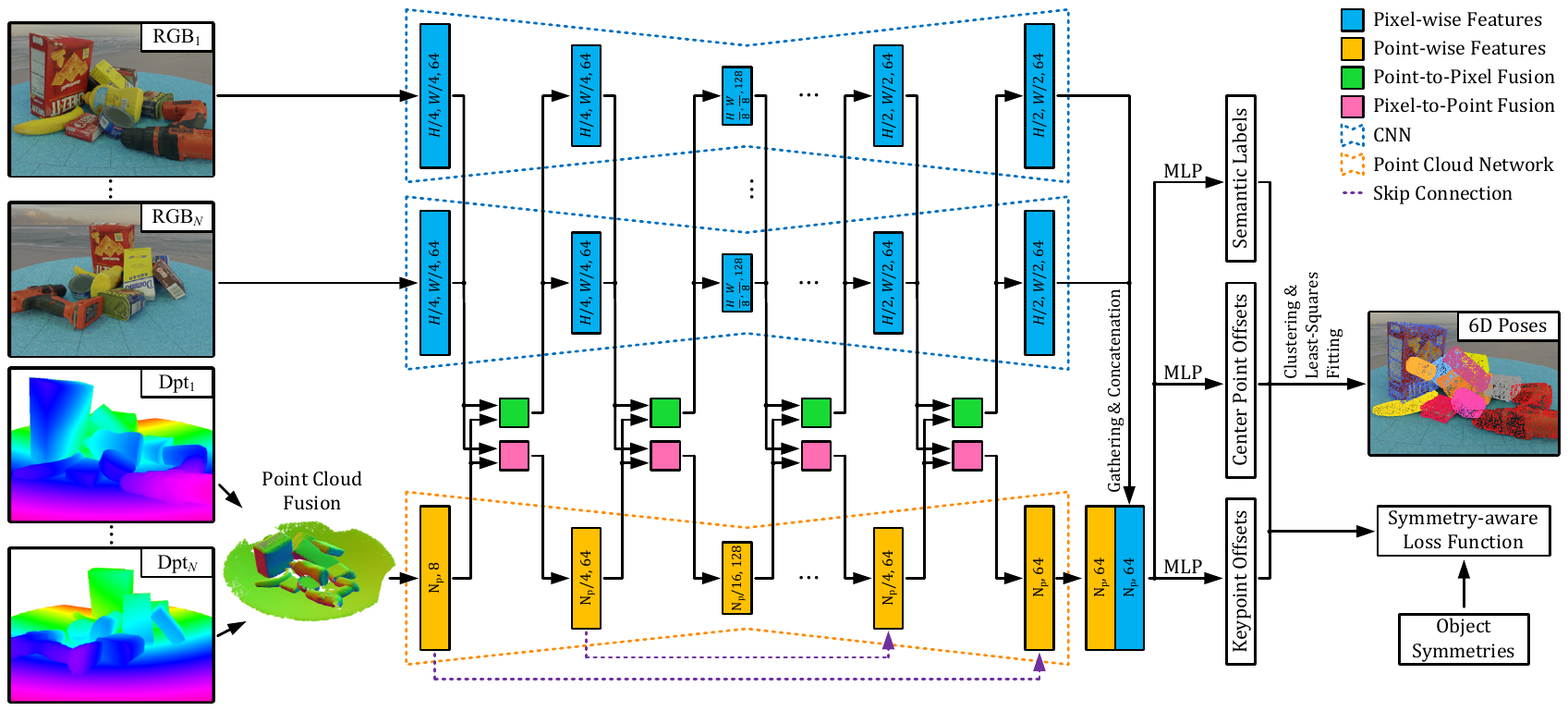}
   \caption{Network architecture of SyMFM6D which fuses $N$ RGB-D input images. Our method converts the $N$ depth images to a single point cloud which is processed by an encoder-decoder point cloud network. The $N$ RGB images are processed by an encoder-decoder CNN. Every hierarchy contains a point-to-pixel fusion module and a pixel-to-point fusion module for deep multi-directional multi-view fusion. We utilize three MLPs with four layers each to regress 3D keypoint offsets, center point offsets, and semantic labels based on the final features. The 6D object poses are computed as in \cite{pvn3d} based on mean shift clustering and least-squares fitting. We train our network by minimizing our proposed symmetry-aware multi-task loss function using precomputed object symmetries. $N_p$ is the number of points in the point cloud. $H$ and $W$ are height and width of the RGB images.}
   \label{fig_architecture}
   \vspace{-2mm}
\end{figure*}

6D object pose estimation describes the task of predicting a rigid transformation $\boldsymbol p = [\boldsymbol R |  \boldsymbol t] \in SE(3)$ which transforms the coordinates of an observed object from the object coordinate system into the camera coordinate system. This transformation is called 6D object pose because it is composed of a 3D rotation $\boldsymbol R \in SO(3)$ and a 3D translation $\boldsymbol t \in \mathbb{R}^3$. 
The designated aim of our approach is to jointly estimate the 6D poses of all objects in a given cluttered scene using multiple RGB-D images which depict the scene from multiple perspectives. We assume the 3D models of the objects and the camera poses to be known as proposed by \cite{mv6d}.

\subsection{Network Overview}

Our symmetry-aware multi-view network consists of three stages which are visualized in \cref{fig_architecture}. 
The first stage receives one or multiple RGB-D images and extracts visual features as well as geometric features which are fused to a joint representation of the scene. 
The second stage performs a detection of predefined 3D keypoints and an instance semantic segmentation.
Based on the keypoints and the information to which object the keypoints belong, we compute the 6D object poses with a least-squares fitting algorithm \cite{leastSquares} in the third stage.

\subsection{Multi-View Feature Extraction}

To efficiently predict keypoints and semantic labels, the first stage of our approach learns a compact representation of the given scene by extracting and merging features from all available RGB-D images in a deep multi-directional fusion manner. For that, we first separate the set of RGB images $\text{RGB}_1, ..., \text{RGB}_N$ from their corresponding depth images $\text{Dpt}_1$, ..., $\text{Dpt}_N$. The $N$ depth images are converted into point clouds, transformed into the coordinate system of the first camera, and merged to a single point cloud using the known camera poses as in \cite{mv6d}. 
Unlike \cite{mv6d}, we employ a point cloud network based on RandLA-Net \cite{hu2020randla} with an encoder-decoder architecture using skip connections.
The point cloud network learns geometric features from the fused point cloud and considers visual features from the multi-directional point-to-pixel fusion modules as described in \cref{sec_multi_view_fusion}.

The $N$ RGB images are independently processed by a CNN with encoder-decoder architecture using the same weights for all $N$ views. The CNN learns visual features while considering geometric features from the multi-directional pixel-to-point fusion modules. We followed \cite{ffb6d} and build the encoder upon a ResNet-34 \cite{resnet} pretrained on ImageNet~\cite{imagenet} and the decoder upon a PSPNet \cite{pspnet}. 

After the encoding and decoding procedures including several multi-view feature fusions, we collect the visual features from each view corresponding to the final geometric feature map and concatenate them. The output is a compact feature tensor containing the relevant information about the entire scene which is used for keypoint detection and instance semantic segmentation as described in \cref{sec_keypoint_detection_and_segmentation}.

\begin{figure*}[tbh]
  \vspace{2mm} 
  \centering  
\begin{subfigure}[b]{0.48\textwidth}
  \includegraphics[page=1, trim = 1mm 6mm 6mm 6mm, clip,  width=1.0\linewidth]{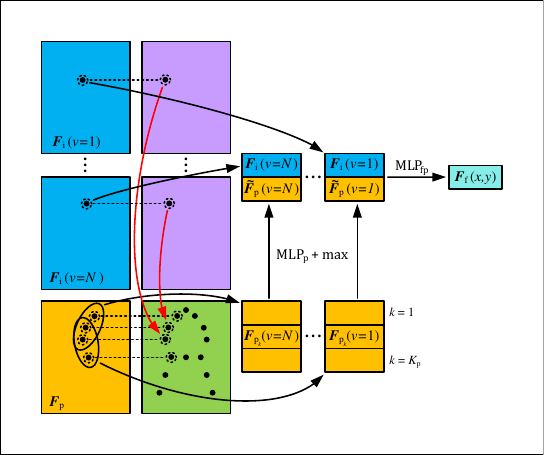}
   \caption{Point-to-pixel fusion module.~~~~}
   \label{fig_pt2px_fusion}
\end{subfigure}
\begin{subfigure}[b]{0.48\textwidth}
  \centering  
  \includegraphics[page=1, trim = 1mm 6mm 6mm 6mm, clip,  width=1.0\linewidth]{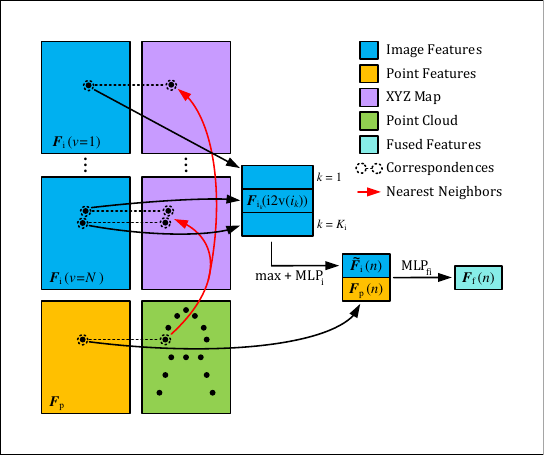}
   \caption{Pixel-to-point fusion module.~~~~~}
   \label{fig_px2pt_fusion}
   \end{subfigure}
      \caption{Overview of our proposed multi-directional multi-view fusion modules. They combine pixel-wise visual features and point-wise geometric features by exploiting the correspondence between pixels and points using the nearest neighbor algorithm. We compute the resulting features using multiple shared MLPs with a single layer and max-pooling.
      For simplification, we depict an example with $N=2$ views and $K_\text{i}=K_\text{p}=3$ nearest neighbors. The points of ellipsis (...) illustrate the generalization for an arbitrary number of views $N$. Please refer to \cite{ffb6d} for better understanding the basic operations.
      }
   \label{fig_fusion_modules}
   \vspace{-1mm}
\end{figure*}

\subsection{Multi-View Feature Fusion}
\label{sec_multi_view_fusion}
In order to efficiently fuse the visual and geometric features from multiple views, we extend the fusion modules of FFB6D~\cite{ffb6d} from bi-directional fusion to \emph{multi-directional fusion}. We present two types of multi-directional fusion modules which are illustrated in \cref{fig_fusion_modules}.
Both types of fusion modules take the pixel-wise visual feature maps and the point-wise geometric feature maps from each view, combine them, and compute a new feature map.
This process requires a correspondence between pixel-wise and point-wise features which we obtain by computing an XYZ map for each RGB feature map based on the depth data of each pixel using the camera intrinsic matrix as in \cite{ffb6d}. To deal with the changing dimensions at different layers, we use the centers of the convolutional kernels as new coordinates of the feature maps and resize the XYZ map to the same size using nearest interpolation as proposed in \cite{ffb6d}.

The \emph{point-to-pixel} fusion module in \cref{fig_pt2px_fusion} computes a 
fused feature map $\bb F_\text{f}$ based on the image features $\bb F_{\text{i}}(v)$ of all views $v \in \{1, \ldots, N\}$.
We collect the $K_\text{p}$ nearest point features $\bb F_{\text{p}_k}(v)$ with $k \in \{1, \ldots, K_\text{p}\}$ from the point cloud for each pixel-wise feature and each view independently by computing the nearest neighbors according to the Euclidean distance in the XYZ map. Subsequently, we process them by a shared MLP before aggregating them by max-pooling, i.e.,
\begin{align} 
    \widetilde{\bb F}_{\text{p}}(v) = \max_{k \in \{1, \ldots, K_\text{p}\}} 
    \Big( \text{MLP}_\text{p}(\bb F_{\text{p}_k}(v)) \Big).
    \label{eq_p2r}
\end{align}
Finally, we apply a second shared MLP to fuse all features $\bb F_\text{i}$ and 
$\widetilde{\bb F}_{\text{p}}$ as 
$\bb F_{\text{f}} = \text{MLP}_\text{fp}(\widetilde{\bb F}_{\text{p}} \oplus \bb F_\text{i})$ where $\oplus$ denotes the concatenate operation.

The \emph{pixel-to-point} fusion module in \cref{fig_px2pt_fusion} collects the $K_\text{i}$ nearest image features $\bb F_{\text{i}_k}(\textrm{i2v}(i_k))$ with $k\in\{1, ..., K_\text{i}\}$. $\textrm{i2v}(i_k)$ is a mapping that maps the index of an image feature to its corresponding view. This procedure is performed for each point feature vector $\bb F_\text{p}(n)$.
We aggregate the collected image features by max-pooling and apply a shared MLP, i.e.,
\begin{align}
    \widetilde{\bb F}_{\text{i}} = \text{MLP}_\text{i} 
    \left( \max_{k \in \{1, \ldots, K_\text{i}\}} 
    \Big( \bb F_{\text{i}_k}(\textrm{i2v}(i_k)) \Big)  
    \right).
    \label{eq_r2p}
\end{align}
One more shared MLP fuses the resulting image features $\widetilde{\bb F}_{\text{i}}$ with the point features $\bb F_\text{p}$ as 
$\bb F_{\text{f}} = \text{MLP}_\text{fi}(\widetilde{\bb F}_{\text{i}} \oplus \bb F_\text{p})$.

\subsection{Keypoint Detection and Segmentation}
\label{sec_keypoint_detection_and_segmentation}
The second stage of our SyMFM6D network contains modules for 3D keypoint detection and instance semantic segmentation following \cite{mv6d}. However, unlike \cite{mv6d}, we use the SIFT-FPS algorithm \cite{lowe1999sift} as proposed by FFB6D \cite{ffb6d} to define eight target keypoints for each object class. SIFT-FPS yields keypoints with salient features which are easier to detect.
Based on the extracted features, we apply two shared MLPs to estimate the translation offsets from each point of the fused point cloud to each target keypoint and to each object center.
We obtain the actual point proposals by adding the translation offsets to the respective points of the fused point cloud. 
Applying the mean shift clustering algorithm \cite{cheng1995meanshift} results in predictions for the keypoints and the object centers.
We employ one more shared MLP 
for estimating the object class of each point in the fused point cloud as in \cite{pvn3d}.

\subsection{6D Pose Computation via Least-Squares Fitting}

Following \cite{pvn3d}, we use the least-squares fitting algorithm \cite{leastSquares} to compute the 6D poses of all objects based on the estimated keypoints. As the $M$ estimated keypoints $\boldsymbol{\widehat{k}}_1, ..., \boldsymbol{\widehat{k}}_M$ are in the coordinate system of the first camera and the target keypoints $\boldsymbol k_1, ..., \boldsymbol k_M$ are in the object coordinate system, least-squares fitting calculates the rotation matrix $\boldsymbol R$ and the translation vector $\boldsymbol t$ of the 6D pose by minimizing the squared loss
\begin{equation}
    L_\text{Least-squares} = \sum_{i=1}^M \norm{\boldsymbol{\widehat{k}_i} - (\boldsymbol R \boldsymbol k_i + \boldsymbol t)}_2^2.
\end{equation}

\subsection{Symmetry-aware Keypoint Detection}

Most related work, including \cite{pvn3d, ffb6d}, and \cite{mv6d} does not specifically consider object symmetries. 
However, symmetries lead to ambiguities in the predicted keypoints as multiple 6D poses can have the same visual and geometric appearance. 
Therefore, we introduce a novel symmetry-aware training procedure for the 3D keypoint detection including a novel symmetry-aware objective function to make the network predicting either the original set of target keypoints for an object or a rotated version of the set corresponding to one object symmetry. Either way, we can still apply the least-squares fitting which efficiently computes an estimate of the target 6D pose or a rotated version corresponding to an object symmetry. To do so, we precompute the set $\boldsymbol{S}_I$ of all rotational symmetric transformations for the given object instance $I$ with a stochastic gradient
descent algorithm \cite{sgdr}.
Given the known mesh of an object and an initial estimate for the symmetry axis, we transform the object mesh along the symmetry axis estimate and optimize the symmetry axis iteratively by minimizing the ADD-S metric \cite{hinterstoisser2012model}.
Reflectional symmetries which can be represented as rotational symmetries are handled as rotational symmetries. 
Other reflectional symmetries are ignored, since the reflection cannot be expressed as an Euclidean transformation.
To consider continuous rotational symmetries, we discretize them into 16 discrete rotational symmetry transformations.

We extend the keypoints loss function of \cite{pvn3d} to become symmetry-aware such that it predicts the keypoints of the closest symmetric transformation, i.e. 
\begin{equation}
    L_\text{kp}(\mathcal{I}) = \frac{1}{N_I} 
    \min_{\boldsymbol{S} \in \boldsymbol{S}_I} 
    \sum_{i \in \mathcal{I}} \sum_{j=1}^M 
    \norm{\boldsymbol{x}_{ij} - \boldsymbol{S}\boldsymbol{\widehat{x}}_{ij}}_2, 
\label{eq_keypoint_loss}
\end{equation}
where $N_I$ is the number of points in the point cloud for object instance $I$, $M$ is the number of target keypoints per object, and $\mathcal{I}$ is the set of all point indices that belong to object instance $I$.  
The vector $\boldsymbol{\widehat{x}}_{ij}$ is the predicted keypoint offset for the $i$-th point and the $j$-th keypoint while $\boldsymbol{x}_{ij}$ is the corresponding ground truth.

\subsection{Objective Function}

We train our network by minimizing the multi-task loss function
\begin{equation}
 \label{eq_total_loss}
    L_\text{multi-task} = \lambda_1 L_\text{kp} 
    + \lambda_2 L_\text{semantic}  
    +  \lambda_3 L_\text{cp},
\end{equation}
where $L_\text{kp}$ is our symmetry-aware keypoint loss from \cref{eq_keypoint_loss}.
$L_\text{cp}$ is an L1 loss for the center point prediction, $L_\text{semantic}$ is a Focal loss \cite{focalLoss} for the instance semantic segmentation, and $\lambda_1=2$, $\lambda_2=1$, and $\lambda_3=1$ are the weights for the individual loss functions as in \cite{ffb6d}.

\section{Experiments}

To demonstrate the performance of our method in comparison to related approaches, we perform extensive experiments on four very challenging datasets.

\subsection{Datasets}

The {\bf YCB-Video dataset} \cite{posecnn} contains a total of 133,827 RGB-D images showing 92 scenes composed of three to nine objects from the 21 Yale-CMU-Berkeley (YCB) objects \cite{ycb}.
Additionally, there are 80,000 synthetic non-sequential \mbox{RGB-D} frames showing a random subset of the YCB objects placed at random positions.

However, most frames from YCB-Video are very similar because they originate from videos with 30 frames per second recorded by a handheld camera that was moved slowly. The videos also do not show the scene from all sides but just from similar perspectives. Furthermore, the scenes do not include strong occlusions, and hence, most object poses are simple to estimate from a single perspective. 
Therefore, we additionally consider the recently proposed photorealistic synthetic datasets {\bf MV-YCB FixCam} and {\bf MV-YCB WiggleCam} \cite{mv6d} as they contain much more difficult scenes with strong occlusions and diverse camera perspectives.
Both datasets depict 8,333 cluttered scenes composed of eleven non-symmetric YCB objects which are randomly arranged so that strong occlusions occur. Each scene is photorealistically rendered from three very different perspectives providing 24,999 RGB-D images with accurate ground truth annotations. Unlike FixCam which uses fixed camera positions while providing accurate camera poses, WiggleCam has varying camera poses which are inaccurately annotated on purpose.

Since FixCam and WiggleCam contain only non-symmetric objects, we created an additional photorealistic synthetic dataset with symmetric and non-symmetric objects called {\bf MV-YCB SymMovCam} using Blender with physically based rendering and domain randomization as in \cite{mv6d}. It also depicts 8,333 cluttered scenes, but they are composed of 8 -- 16 objects randomly chosen from the 21 YCB objects which results in very strong occlusions. For each scene, we created four cameras at changing positions around the scene with the restriction that in each quadrant there is only one camera so that the perspectives are very distinct. This results in a total of 33,332 annotated RGB-D images.

\subsection{Training Procedure}

For training our model in single-view mode on YCB-Video, we randomly use the synthetic and real images of YCB-Video with a ratio of 4:1. Since consecutive real frames are very similar, we consider only every seventh real frame. For training a multi-view model, we start from the corresponding single-view checkpoint and continue training with batches of real YCB-Video frames. 
For training on FixCam and WiggleCam we follow \cite{mv6d} and use random permutations of the three available camera views. For SymMovCam, we take a random subset of three views from the available four views.

\subsection{Evaluation Metrics}
\label{sec_eval_metrics}

We evaluated our method using the area-under-curve (AUC) metrics for ADD-S and \mbox{ADD(-S)} and the precision metrics ADD-S \textless ~\SI{2}{cm} and \mbox{ADD(-S)} \textless ~\SI{2}{cm} as these metrics are most commonly used in related work \cite{cosypose, ffb6d, mv6d}.

\subsection{Baseline Methods}

We compare our methods with many established and some very recent methods namely 
DenseFusion \cite{densefusion}, CosyPose \cite{cosypose}, PVN3D \cite{pvn3d}, FFB6D \cite{ffb6d}, ES6D \cite{es6d}, and MV6D \cite{mv6d}.

\subsection{Results on YCB-Video}

\begin{table}[b]
    \centering
\begin{tabular}{l|ccc}
    \toprule
                           &    ADD-S  &    ADD(-S) \\\midrule
DenseFusion (per-pixel)    &     91.2  &     82.9   \\ 
DenseFusion (iterative)    &     93.2  &     86.1   \\
CosyPose                   &     89.8  &     84.5   \\
PVN3D                      &     95.5  &     91.8   \\     
FFB6D                      &     96.6  &     92.7   \\     
ES6D                       &     93.6  &     89.0   \\   
SyMFM6D                    & \tb{96.8} & \tb{94.1}  \\ 
\bottomrule
\end{tabular}
    \caption{Single-view results on YCB-Video using the AUC metrics for ADD-S and \mbox{ADD(-S)}. The best results are printed in bold.}
    \label{tab_ycbv_sv}
\end{table}

\cref{tab_ycbv_sv} compares the single-view performance of our SyMFM6D network with all baseline methods using the AUC of ADD-S and \mbox{ADD(-S)} on YCB-Video. Please note that MV6D corresponds to PVN6D in the single-view scenario. The results show that our approach copes very well with the dynamic camera setup of YCB-Video while outperforming all methods significantly. On the symmetry-aware \mbox{ADD(-S)} AUC metric, SyMFM6D outperforms the current state-of-the-art FFB6D by even \SI{1.5}{\%}. 
Please note that unlike DenseFusion (iterative) and CosyPose, our approach does not perform computationally expensive post processing or iterative refinement procedures.

To examine the effect of our symmetry-aware training procedure, we provide an object-wise evaluation of the three best single-view methods on YCB-Video in \cref{fig_ycb_sv_objects}. Please note that in single-view mode, our model architecture is the same as FFB6D except for our novel symmetry-aware loss function. 
The results show that not only most symmetric objects (highlighted in bold) are estimated more accurate but also most non-symmetric objects.
This indicates that there is a synergy effect which improves the keypoint detection for non-symmetric objects due to an improvement of the keypoint detection for symmetric objects.

\begin{table}[tbp]
    \vspace{2mm}
    \centering
    \begin{tabular}{l|ccc}
        \toprule 
        Object class  		   &   PVN3D  &   FFB6D  &  SyMFM6D  \\\midrule
        Master chef can        &    80.5  &    80.6  &\tb{80.7} \\
        Cracker box            &    94.8  &    94.6  &\tb{94.9} \\
        Sugar box              &    96.3  &\tb{96.6} &\tb{96.6} \\
        Tomato soup can        &    88.5  &\tb{89.6} &    87.9  \\
        Mustard bottle         &    96.2  &    97.0  &\tb{97.8} \\
        Tuna fish can          &    89.3  &    88.9  &\tb{92.3} \\
        Pudding box            &\tb{95.7} &    94.6  &    93.3  \\
        Gelatin box            &    96.1  &\tb{96.9} &    96.1  \\
        Potted meat can        &    88.6  &    88.1  &\tb{90.0} \\
        Banana                 &    93.7  &    94.9  &\tb{95.2} \\
        Pitcher base           &    96.5  &    96.9  &\tb{97.5} \\
        Bleach cleanser        &    93.2  &\tb{94.8} &    93.9  \\
        \tb{Bowl}              &    90.2  &    96.3  &\tb{96.4} \\
        Mug                    &    95.4  &    94.2  &\tb{95.7} \\
        Power drill            &    95.1  &    95.9  &\tb{96.4} \\
        \tb{Wood block}        &    90.4  &    92.6  &\tb{95.2} \\
        Scissors               &    92.7  &    95.7  &\tb{95.8} \\
        Large marker           &\tb{91.8} &    89.1  &    90.0  \\
        \tb{Large clamp}       &    93.6  &    96.8  &\tb{96.9} \\
        \tb{Extra large clamp} &    88.4  &\tb{96.0} &    95.3  \\
        \tb{Foam brick}        &    96.8  &    97.3  &\tb{97.6} \\\midrule
        ALL                    &    91.8  &    92.7  &\tb{94.1} \\\bottomrule
    \end{tabular} 
	\caption{Single-view results on YCB-Video evaluated for each object class individually using the \mbox{ADD(-S)} AUC metric. Symmetric objects and the best results are printed in bold.}
	\label{fig_ycb_sv_objects}
\end{table}

\begin{figure*}[htbp]
        \vspace{2mm}
	\centering
	\begin{minipage}{0.24\textwidth}
		\centering
		\textbf{Original View}
	\end{minipage}%
	\begin{minipage}{0.24\textwidth}
		\centering
		\textbf{Ground Truth}
	\end{minipage}%
	\begin{minipage}{0.24\textwidth}
		\centering
		\textbf{FFB6D}
	\end{minipage}%
	\begin{minipage}{0.24\textwidth}
		\centering
		\textbf{SyMFM6D}
	\end{minipage}% 
	\setkeys{Gin}{width=0.24\linewidth}
	\includegraphics{figures/ycb_sv/0055_001042_rgb_1800.png}\,%
	\includegraphics{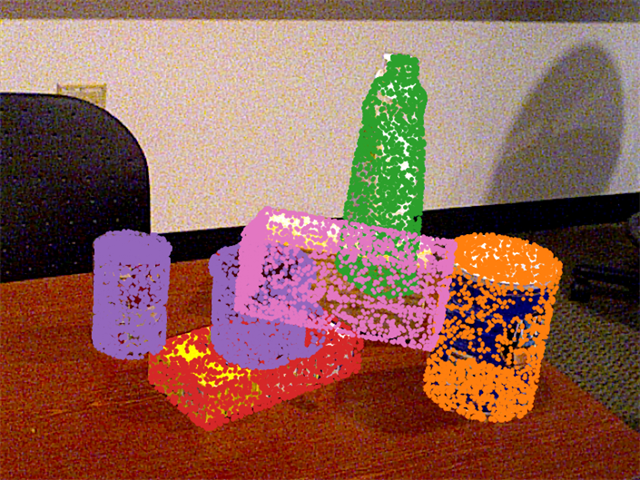}\,%
	\includegraphics{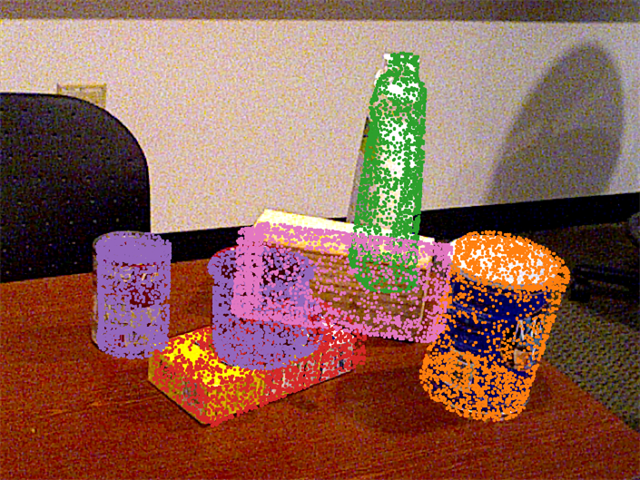}\,%
	\includegraphics{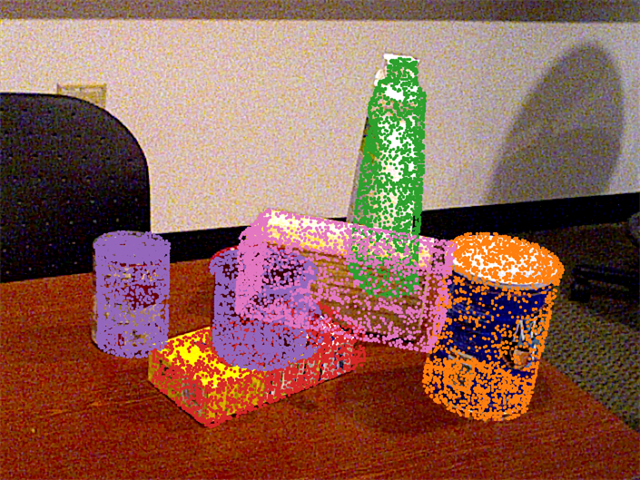}\,%
	\vspace{0.7mm}
	
	\includegraphics{figures/ycb_sv/0051_000636_rgb_864.png}\,%
	\includegraphics{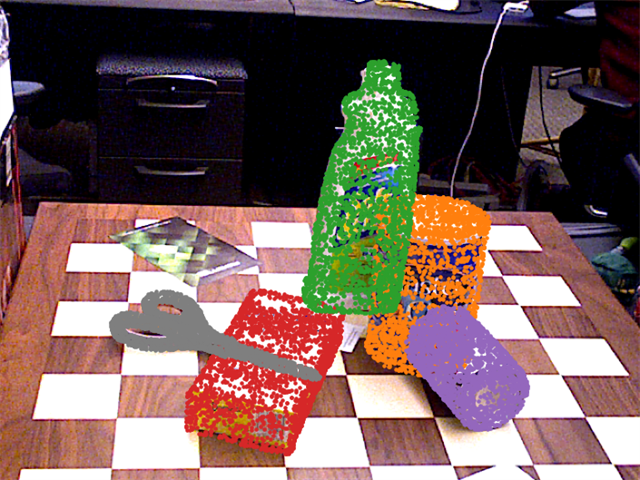}\,%
	\includegraphics{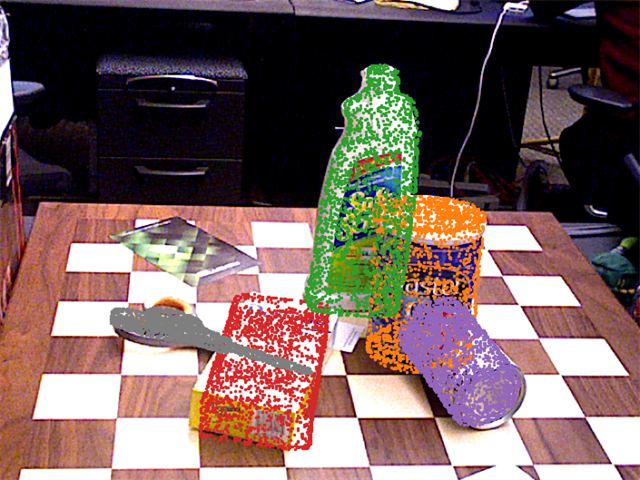}\,%
	\includegraphics{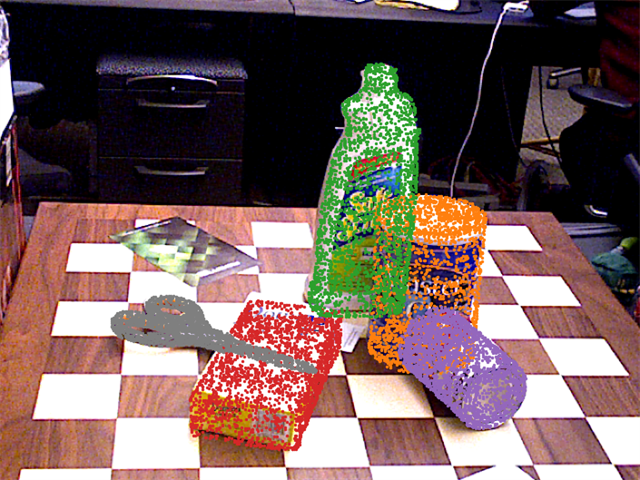}\,%
	\vspace{0.7mm}
	
	\includegraphics{figures/ycb_sv/0058_000553_rgb_2461.png}\,%
	\includegraphics{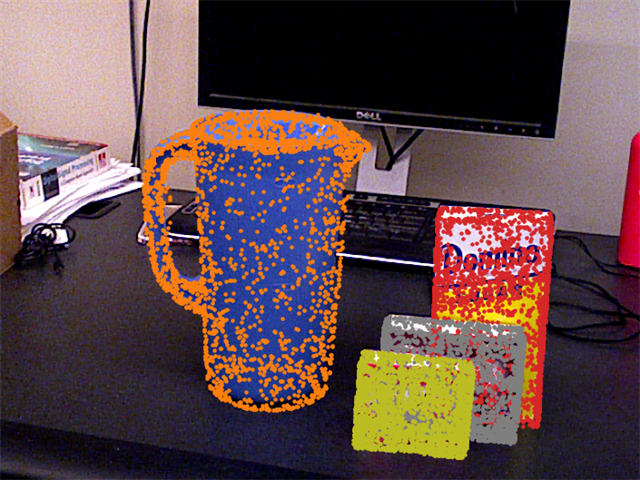}\,%
	\includegraphics{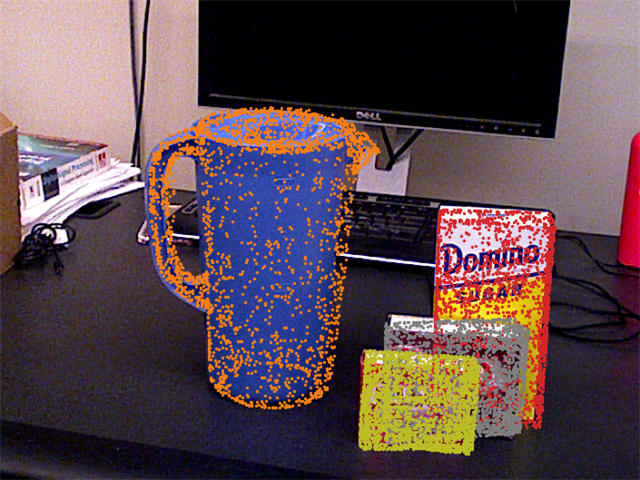}\,%
	\includegraphics{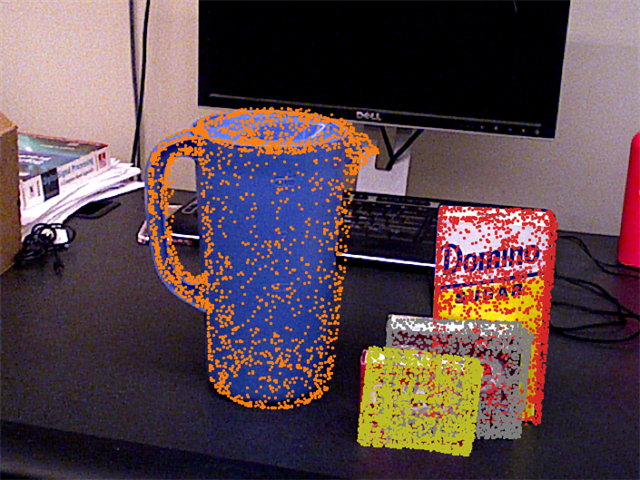}\,%
	\caption{Comparison of 6D pose predictions on single frames of the YCB-Video dataset.}
	\label{fig_ycbv_sv}
\end{figure*}

\cref{fig_ycbv_sv} shows a visualization of three scenes of YCB-Video with 6D pose ground truth, predictions of FFB6D, and predictions of our SyMFM6D network using only the depicted view. It can be seen that both FFB6D and SyMFM6D estimate very accurate poses as the scenes of YCB-Video contain only a few objects and not many occlusions. However, SyMFM6D predicts even more accurate poses than FFB6D due to our proposed symmetry-aware training procedure.

\cref{tab_ycbv_mv} compares our multi-view results with all multi-view baseline methods on YCB-Video using three and five input views.
We see that our approach with disabled symmetry training procedure already outperforms all previous multi-view methods significantly. Enabling the symmetry awareness further improves the results slightly. However, 
using more views does not improve the accuracy as most views of YCB-Video are very similar in which case additional views do not provide beneficial information while the learning problem of fusing different views becomes slightly harder.

\begin{table}[!hbt]
    \tabcolsep=1.35mm
    \centering
\begin{tabular}{l|cc|cc}
    \toprule 
                  & \multicolumn{2}{c|}{ADD-S}
                                          & \multicolumn{2}{c}{ADD(-S)} \\
                  &  3 views  &  5 views  &  3 views  &  5 views  \\\midrule
CosyPose          &     92.3  &     93.4  &     87.7  &     88.8  \\
MV6D              &     91.2  &     91.1  &     85.6  &     84.0  \\
SyMFM6D (no sym)  &     95.2  &     95.2  &     91.5  &     91.4  \\
SyMFM6D           & \tb{95.4} & \tb{95.4} & \tb{91.7} & \tb{91.6} \\
\bottomrule
\end{tabular}
    \caption{Quantitative multi-view results on YCB-Video. The best results are printed in bold.}
    \label{tab_ycbv_mv}
\end{table}

\subsection{Results on MV-YCB FixCam, WiggleCam and SymMovCam}

We show the quantitative results on the datasets MV-YCB FixCam, MV-YCB WiggleCam, and MV-YCB SymMovCam in \cref{tab_fixCam_wiggleCam}. It includes a comparison with two modified CosyPose (CP) versions with and without known camera poses as presented by \cite{mv6d}.
Our SyMFM6D network yields the best results on all metrics on all three datasets. This shows that SyMFM6D copes very well with the strong occlusions in the datasets. The results on WiggleCam are just slightly worse than on FixCam which demonstrates that our approach is robust towards inaccurately known camera poses.

On the novel SymMovCam dataset, our method outperforms the baselines by a much larger margin than on FixCam and WiggleCam. This is due to the symmetric objects in the datasets on which the keypoint estimation of the baseline methods is inaccurate. The results also prove that our approach is robust to very dynamic camera setups where the cameras are mounted at varying positions.

\begin{table*}[h]
	\tabcolsep=1.0mm
	\centering
	\begin{tabular}{r|cccccc|cccccc|ccccc}
    \toprule
                           &     \multicolumn{6}{c|}{MV-YCB FixCam}                        &      \multicolumn{6}{c|}{MV-YCB WiggleCam}                    &      \multicolumn{5}{c}{MV-YCB SymMovCam}            \\
                           &  PVN3D   &   FFB6D  &   CP   &    CP    &   MV6D   &   Ours   &  PVN3D   & FFB6D    &   CP   &   CP     &  MV6D    &  Ours    &  PVN3D   & FFB6D    &   Ours   &    MV6D  &  Ours    \\ 
    Number of views        &   1      &     1    &   3    &    3     &   3      &    3     &    1     & 1        &   3    &   3      &   3      &    3     &    1     &    1     &     1    &     3    &    3     \\
    Known cam poses        &\checkmark&\checkmark&$\times$&\checkmark&\checkmark&\checkmark&\checkmark&\checkmark&$\times$&\checkmark&\checkmark&\checkmark&\checkmark&\checkmark&\checkmark&\checkmark&\checkmark\\
    \midrule                                                                                      
    ADD-S AUC              &  81.3    &   82.3   &  90.8  &   91.9   &  96.9    &\tb{97.3} &   80.8   &   81.9   & 90.0   &  91.3    &    96.2  &\tb{96.7} &   75.0   &   79.9   &   80.6   &   92.8   & \tb{94.2}\\
    ADD(-S) AUC            &  74.9    &   76.3   &  82.4  &   84.6   &  94.8    &\tb{95.6} &   74.0   &   75.5   & 81.0   &  83.4    &    93.0  &\tb{94.2} &   68.5   &   75.6   &   76.7   &   88.7   & \tb{91.6}\\
    ADD-S \textless   ~\SI{2}{cm} &  82.1    &   83.6   &  92.9  &   93.0   &  98.8    &\tb{98.9} &   82.0   &   83.4   & 92.3   &  92.6    &    98.7  &\tb{98.8} &   77.2   &   81.1   &   81.9   &   96.3   & \tb{96.6}\\
    ADD(-S) \textless ~\SI{2}{cm} &  73.0    &   74.8   &  80.6  &   82.4   &  96.5    &\tb{96.8} &   72.4   &   74.0   & 78.9   &  81.6    &\tb{96.0} &\tb{96.0} &   64.5   &   74.5   &   76.3   &   91.6   & \tb{93.6}\\
    \bottomrule
	\end{tabular}
	\caption{Quantitative results on the datasets MV-YCB FixCam (left), MV-YCB WiggleCam (middle), and MV-YCB SymMovCam (right). The baseline CosyPose (CP) uses PVN3D as backend network as described in \cite{mv6d}. The best results per dataset are printed in bold.}
	\label{tab_fixCam_wiggleCam}
\end{table*}

\subsection{Keypoint Visualization}

\begin{figure}[!tbh]
  \centering
\begin{subfigure}[b]{0.49\columnwidth}
  \includegraphics[width=1.0\columnwidth]{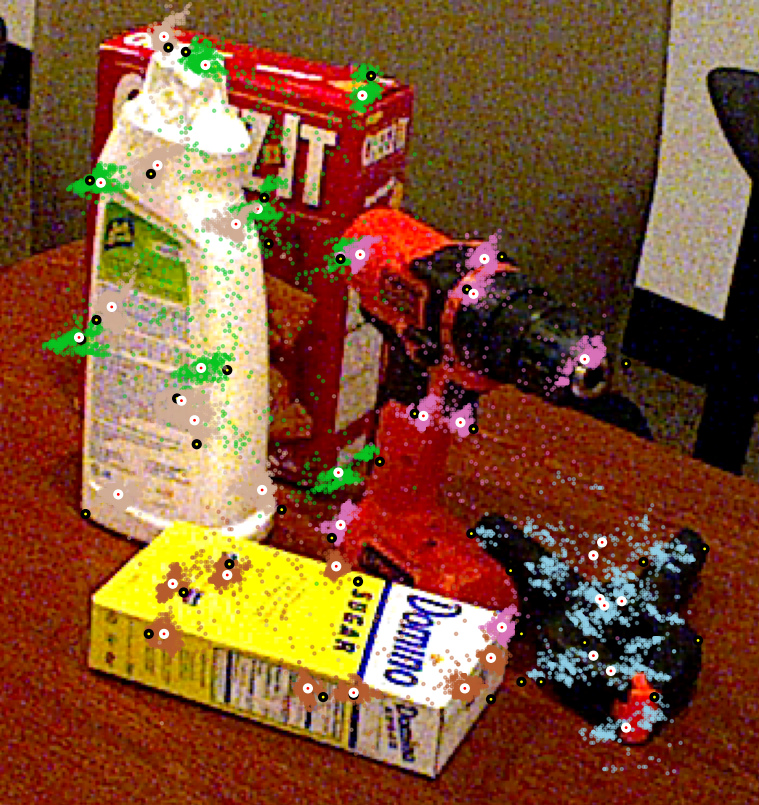}
   \caption{FFB6D}
   \label{fig_ycbv_sv_keypoints_FFB6D}
\end{subfigure}
\begin{subfigure}[b]{0.49\columnwidth}
  \centering
  \includegraphics[width=1.0\columnwidth]{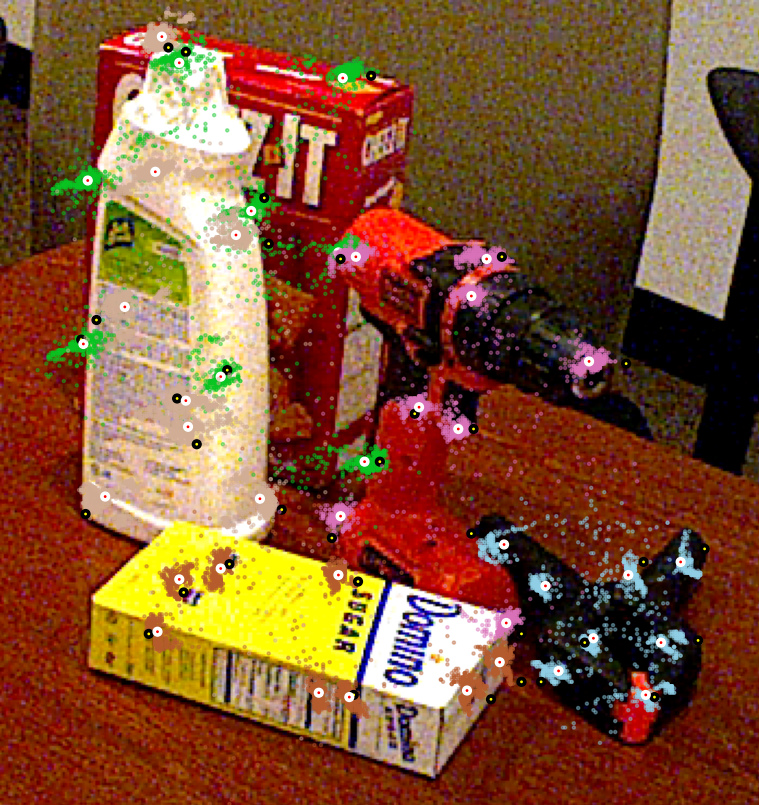}
   \caption{SyMFM6D}
   \label{fig_ycbv_sv_keypoints_SyMFM6D}
   \end{subfigure}
	\caption{Visualization of the predicted keypoints on single frames of the YCB-Video dataset.} 
   \label{fig_ycbv_sv_keypoints}
\end{figure}

\cref{fig_ycbv_sv_keypoints} shows predicted keypoints of FFB6D and SyMFM6D in a YCB-Video scene. We additionally visualize the keypoint proposals of each object in individual colors.
The resulting predicted keypoints are white, the target keypoints are black. You can see that both FFB6D and SyMFM6D predict very accurate keypoints on all non-symmetric objects. However, FFB6D fails to predict accurate keypoints on the large clamp which has one discrete rotational symmetry. This shortcoming of FFB6D is also apparent on other symmetric objects. We believe that this is caused by the ambiguities of the object poses resulting in ambiguous target keypoints which results in averaging over the multiple solutions given by the symmetry. Therefore, the training loss is minimized when predicting keypoints on the symmetric axis rather than predicting them on the desired target locations. SyMFM6D in contrast overcomes this problem by our novel symmetry-aware training procedure as it can be seen in \cref{fig_ycbv_sv_keypoints_SyMFM6D}.

\subsection{Implementation Details and Runtime}

We trained our network up to seven days on four NVIDIA Tesla V100 GPUs with \SI{32}{GB} of memory. 
The network architecture of our SyMFM6D approach has 3.5 million trainable parameters and requires about \SI{46}{ms} for processing a single RGB-D image on a single GPU. 
Mean shift clustering and least-squares fitting for computing a 6D pose require additional \SI{14}{ms} per object. 
Please visit our previously mentioned GitHub repository for code, datasets, and further details.

\section{Conclusion}

In this work, we present SyMFM6D, a novel approach for symmetry-aware multi-view 6D object pose estimation based on a deep multi-directional fusion network for RGB-D data. We additionally propose a novel method for predicting predefined 3D keypoints of symmetric objects based on a symmetry-aware objective function. Using the 3D keypoint predictions and an instance semantic segmentation, we compute the 6D poses of all objects in the scene simultaneously with least-squares fitting. Our experiments show that our symmetry-aware training procedure significantly improves the 6D pose estimation accuracy of both symmetric and non-symmetric objects due to synergy effects. Our method outperforms the state-of-the-art in single-view and multi-view 6D pose estimation on four very challenging datasets. We furthermore demonstrate the robustness of our approach towards inaccurately known camera poses and dynamic camera setups.

{\small
\bibliographystyle{IEEEtran}
\bibliography{IEEEabrv}
}

\end{document}